\documentclass[10pt,twocolumn,letterpaper]{article}

\usepackage{iccv}
\usepackage{times}
\usepackage{epsfig}
\usepackage{graphicx}
\usepackage{amsmath}
\usepackage{amssymb}

\usepackage[utf8]{inputenc} %
\usepackage[T1]{fontenc}    %
\usepackage{url}            %
\usepackage{booktabs}       %
\usepackage{amsfonts}       %
\usepackage{nicefrac}       %
\usepackage{microtype}      %

\usepackage{algorithm}
\usepackage{algpseudocode}
\usepackage{caption}
\usepackage{subcaption}
\usepackage{wrapfig}

\def\Eqref#1{Eq.(\ref{#1})}
\DeclareMathOperator*{\argmax}{arg\,max}
\newcommand{\boldspacepar}[1]{\par\noindent\textbf{#1}}

\usepackage{pifont}%
\newcommand{\xmark}{\ding{55}}%
\usepackage{multirow}

\newcommand{\review}[1]{#1}

\usepackage[pagebackref=true,breaklinks=true,letterpaper=true,colorlinks,bookmarks=false]{hyperref}

\iccvfinalcopy %

\def\iccvPaperID{1108} %

\ificcvfinal\fi

\begin{document}

\title{Continual Prototype Evolution:\\ Learning Online from Non-Stationary Data Streams}

\author{Matthias De Lange \& Tinne Tuytelaars\\
KU Leuven\\
{\tt\small \{matthias.delange,tinne.tuytelaars\}@kuleuven.be}
}

\maketitle
\ificcvfinal\thispagestyle{empty}\fi

\begin{abstract}
   Attaining prototypical features to represent class distributions is well established in representation learning.
    However, learning prototypes online from streaming data proves a challenging endeavor as they rapidly become outdated, caused by an ever-changing parameter space during the learning process.
    Additionally, continual learning does not assume the data stream to be stationary, typically resulting in catastrophic forgetting of previous knowledge.
    As a first, we introduce a system addressing both problems, where prototypes evolve continually in a shared latent space, enabling learning and prediction at any point in time.
    In contrast to the major body of work in continual learning, data streams are processed in an online fashion, without additional task-information, and an efficient memory scheme provides robustness to imbalanced data streams.
    Besides nearest neighbor based prediction, learning is facilitated by a novel objective function, encouraging cluster density about the class prototype and increased inter-class variance. Furthermore, the latent space quality is elevated by pseudo-prototypes in each batch, constituted by replay of exemplars from memory. 
    As an additional contribution, we generalize the existing paradigms in continual learning to incorporate data incremental learning from data streams by formalizing a two-agent learner-evaluator framework.
    We obtain state-of-the-art performance by a significant margin on eight benchmarks, including three highly imbalanced data streams.\footnote{ Code: \scriptsize{\texttt{github.com/Mattdl/ContinualPrototypeEvolution}}}%
\end{abstract}

\begin{figure}[!ht]
  \centering
  \includegraphics[clip,trim={0cm 0cm 0cm 0cm},width=\linewidth]{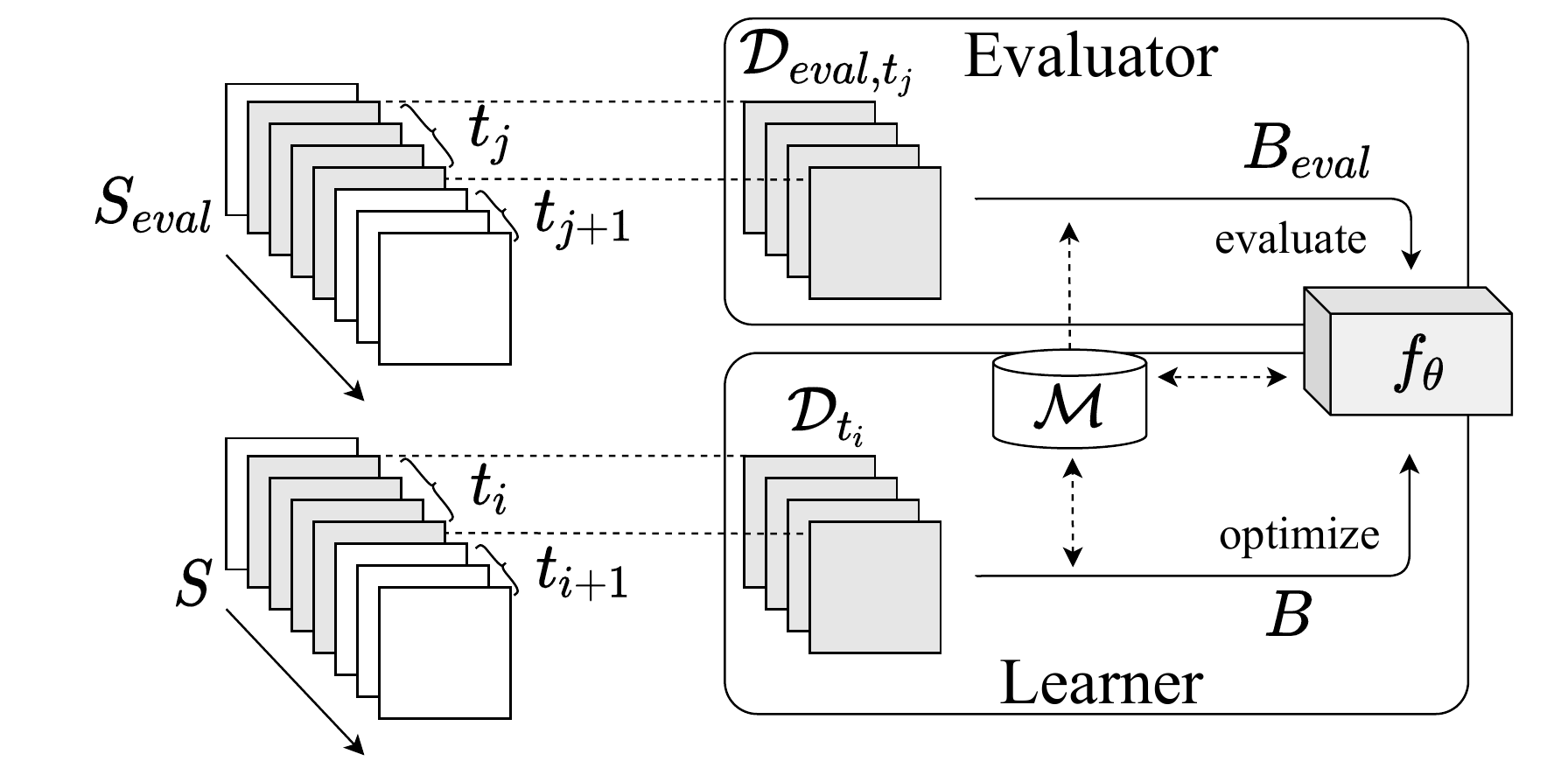}
    \caption{\label{fig:framework}
       Overview of the \emph{learner-evaluator} framework, overcoming the static training and testing paradigms by explicitly modelling continual optimization and evaluation from data streams in the \emph{learner} and \emph{evaluator} agents.
       The framework generalizes to both continual learning and concept drift with resources transparently defined as the horizon $\mathcal{D}$ and operational memory $\mathcal{M}$.
}
\end{figure}

\section{Introduction}
The prevalence of data streams in contemporary applications urges systems to learn in a continual fashion. Autonomous vehicles, sensory robot data, and video streaming yield never-ending streams of data, with abrupt changes in the observed environment behind every vehicle turn, robot entering a new room, or camera cut to a subsequent scene.
 Alas, learning from streaming data is far from trivial due to these changes, as learning schemes for neural networks have been developed assuming identically and independently distributed~(iid) data. When this assumption is violated, catastrophic forgetting of previously acquired knowledge occurs \cite{french1999catastrophic}, emanating a trade-off between neural stability to retain the current state of knowledge and neural plasticity to swiftly adopt the new knowledge \cite{GrossbergStephen1982Soma}. The search for this delicate balance is considered the main hurdle for continual learning systems.

Although a lot of progress has been established in the literature, often strong assumptions apply~\cite{de2019continual}, impeding applicability for real-world systems. Static training and testing paradigms prevail, whereas a true continual learner should enable both training and testing simultaneously and independently. Therefore, we propose the two-agent \emph{learner-evaluator} framework, redefining perspective on existing paradigms in the field. Within this framework, we introduce \emph{data incremental learning}, enabling 
completely task-free learning and evaluation.

Furthermore, we introduce \emph{Continual Prototype Evolution} (CoPE), a new online data incremental learner with prototypes perpetually representing the most salient features of the class population, shifting the catastrophic forgetting problem from the full network parameter space to the lower-dimensional latent space.
As a first, our prototypes evolve continually with the data stream, enabling learning and evaluation at any point in time.
Similar to representativeness heuristics in human cognition~\cite{kahneman1972subjective},  the class prototypes are the cornerstone for nearest neighbor classification.
Additionally, our system is robust to highly imbalanced data streams by the combination of replay with a balancing memory population scheme.
We find batch information in the latent space to be crucial in the challenging non-stationary and online processing regime, which we incorporate in our novel \emph{pseudo-prototypical proxy loss}.

\section{The learner-evaluator framework}
\label{sec:framework}

To date, the paradigms of task, class, and domain incremental learning \cite{van2018generative} dominate the continual learning literature.
However, strong and differing assumptions often lead to confusion and overlap between implementations of these definitions.
Furthermore, the concept of a static training and testing phase is still ubiquitous, whereas continual learning systems should enable both aspects continually and independently.
Therefore, we propose a generalizing framework which disentangles the continual learning system into two agents: the \emph{learner} and the \emph{evaluator}.
Figure~\ref{fig:framework} presents an overview of the framework.

The learning agent learns predicting function ${f_{\theta}: \mathcal{X} \rightarrow \mathcal{Y}}$ parameterized by $\theta$, mapping the input space $\mathcal{X}$ to the target output space $\mathcal{Y}$. The learner receives data samples $\left({\bf x}_i, {\bf y}_i\right)$ from stream $S$ and has simultaneous access to the \emph{horizon} $\mathcal{D}$, i.e.~the observable subset of stream $S$ which can be processed for multiple iterations. 
Data sample $i$ is constituted by input feature ${\bf x}_i \in \mathcal{X}$ and corresponding (self-)supervision signal ${\bf y}_i$ for which the output space for classification is defined as a discrete set of observed classes $\mathcal{Y}_{i} \leftarrow \mathcal{Y}_{i-1} \cup \left\{ {\bf y}_i \right\}$.
To manage memory usage and to enable multiple updates and stochasticity in the optimization process, updates for $\theta$ are typically performed based on a small-scale \emph{processing batch} $B \subseteq \mathcal{D}$. 
The data and size of the horizon $\mathcal{D}$ are determined by the specific setup or application, ranging from standard offline learning with $\mathcal{D}=S$ to online continual learning with  $\mathcal{D} = B$.
 Furthermore, the learner might need additional resources after observing data from $B \subseteq \mathcal{D}$, such as stored samples or model copies, confined by the \emph{operational memory} $\mathcal{M}$.

The evaluating agent acts independently from the learner by evaluating $f_{\theta}$ with horizon $\mathcal{D}_{eval}$ from the evaluation stream $S_{eval}$, with small-scale processing batches $B_{eval} \subseteq \mathcal{D}_{eval}$.
This stream can contain yet unobserved concepts by the learner in $S$ to measure zero-shot performance.
The framework provides leeway for the concept distributions in $S_{eval}$ being either static or dynamically evolving, determining how performance of the learner is measured.
On the one hand, static concept distributions can measure the degree to which the knowledge of learned concepts is preserved, as commonly used in continual learning.
On the other hand, evolving concept distributions measure performance for the current distribution in horizon $\mathcal{D}_{eval}$ only, where concepts might drift from their original representation, also known as concept drift~\cite{schlimmer1986beyond}.
Evaluation can occur asynchronously on-demand or periodically with periodicity $\rho$ determining the resolution of the evaluation samples.

\boldspacepar{Task, class, and domain incremental learning} are based on the composition of the learner's observable stream subset in horizon $\mathcal{D}_t$.
In all three scenarios, the learner receives data of the form  $\left({\bf x}_i, {\bf y}_i, t_i\right)$ with 
$t_i$ the indicator for horizon $\mathcal{D}_{t=t_i}$ of the current task, set of classes, or domain.
In \emph{task incremental learning} the learner's horizon spans all data of a given task, i.e. $\mathcal{D}_t=\left\{ \left(  {\bf x}_i, {\bf y}_i, t_i\right) \in S \ | \ t=t_i \right\}$ \cite{de2019continual,van2019three}.
Moreover, this setup assumes the evaluator to receive data in the same form as the learner, hence having explicit access to $t_i$. This confines prediction to a task-specific subset of output nodes, also referred to as a head.
In \emph{class incremental learning}, the horizon is determined based on a subset of classes $\mathcal{Y}_t$, i.e. $\mathcal{D}_t=\left\{ \left(  {\bf x}_i, {\bf y}_i, t_i\right) \in S \ | \ {\bf y}_i \in \mathcal{Y}_{t=t_i} \right\}$, for which the learner requires $t_i$ to identify the transition between subsequent class subsets $\mathcal{Y}_{t_i}$ in $S$ \cite{rebuffi2017icarl,castro2018end,shmelkov2017incremental,wu2018incremental}.
However, the evaluator considers the entire output space without the need for identifier $t$.
\emph{Domain incremental learning} holds the same assumptions as class incremental learning, although with concepts drifting from one domain to the other for a typically fixed output space, exemplified by the widely used permuted-MNIST setup \cite{goodfellow2013empirical}.
To measure to what extend knowledge is preserved, continual learning assumes in general that for new $t_i$'s a fixed set of evaluation samples is added to the evaluator's horizon $\mathcal{D}_{eval,t=t_i}$, drawn mutually exclusively from the same distribution as the learner's horizon $\mathcal{D}_{t=t_i}$.

\boldspacepar{Data incremental learning} is a more general paradigm we introduce to facilitate learning from any data stream, with no assumption on the horizon or the order of observing the data.
In contrast to existing paradigms, when the \emph{learner} observes horizon $\mathcal{D}$ of data stream $S$, data incremental learning does not disclose an identifier $t$.
Consequently, there is no explicit indication to which subset of the stream is being observed in the horizon $\mathcal{D}$.
Therefore, the learner either processes observed data directly in an online fashion with processing batch $B = \mathcal{D}$, or infers an implicit identifier $t$ from statistics in stream $S$.
Similar to class and domain incremental learning, the \emph{evaluator} operates without $t$ on the full output space.
The data incremental learning paradigm endows continual learning systems with increased practical use, as real-world streaming applications typically lack supervision signal $t$.
Moreover, making this identifier available may depend on arbitrary choices, such as how to delineate tasks, introducing unwanted bias.

\review{
Data incremental learning resembles online learning \cite{shalev2011online} in its independence of identifier $t$. However, online learning makes the rigorous assumption of learning from an iid data stream, whereas data incremental learning 
disregards this assumption, and additionally generalizes beyond the online learning horizon ($\mathcal{D} = B$), by allowing to store larger subsets of the data stream ($\mathcal{D} \subset S$).
The data incremental learning paradigm is also referred to as task-free \cite{aljundi2019task} or task-agnostic \cite{zeno2018task} learning, but is here defined based on transition $t$ of the horizon, which is more general than the task-based definition generating possible ambiguity of how a task is specifically defined.
Note that online continual learning mostly refers to the four continual learning paradigms with horizon $\mathcal{D} = B$, which can also be viewed as learning for a single epoch for each task, class subset, or domain.
Table~\ref{tab:learner-eval} compares data incremental learning with online learning and the three main continual learning paradigms.
}

\begin{table}[]
\centering
\resizebox{\linewidth}{!}{%
\begin{tabular}{@{}llllll@{}}
\toprule
                         & \multicolumn{1}{c}{\textbf{evaluator}}   & \multicolumn{3}{c}{\textbf{learner}}                                                                                             \\ \cmidrule(lr){2-2} \cmidrule(lr){3-5}
                         & \multicolumn{1}{l}{\textit{sample}}      & \multicolumn{1}{l}{\textit{sample}}      & \multicolumn{1}{l}{\textit{horizon $\mathcal{D}$}} &  \multicolumn{1}{c}{\textit{iid}}  \\ \midrule
\textbf{online learning} & $\left({\bf x}_i, {\bf y}_i\right)$      & $\left({\bf x}_i, {\bf y}_i\right)$          & batch ($\mathcal{D} = B$)           & \checkmark   \\ \addlinespace
\textbf{continual learning} &                                          &                                          &                                  &                                                    \\ 
\text{task incr.}      & $\left({\bf x}_i, {\bf y}_i, t_i\right)$ & $\left({\bf x}_i, {\bf y}_i, t_i\right)$  & task ($\mathcal{D}_{t=t_i}$)        & \xmark       \\
\text{class incr.}     & $\left({\bf x}_i, {\bf y}_i\right)$ & $\left({\bf x}_i, {\bf y}_i,  t_i\right)$          & class subset ($\mathcal{D}_{t=t_i}$)    & \xmark       \\
\text{domain incr.}    & $\left({\bf x}_i, {\bf y}_i\right)$ & $\left({\bf x}_i, {\bf y}_i,  t_i\right)$          & domain ($\mathcal{D}_{t=t_i}$)     & \xmark       \\
\text{data incr.}      & $\left({\bf x}_i, {\bf y}_i\right)$      & $\left({\bf x}_i, {\bf y}_i\right)$          & any subset ($B \leq \mathcal{D} < S$)      & \xmark       \\ \addlinespace

\bottomrule
\end{tabular}%
}
\caption{\label{tab:learner-eval} 
\review{The learner-evaluator framework disentangling online learning with four continual learning paradigms. Data incremental learning is fully independent of identifier $t$ as in online learning, but evades the iid assumption.
}
}
\vspace{-0.25cm}
\label{tab:my-table}
\end{table}

\section{Prior work}
\label{sec:relwork}

Continual learning systems are able to learn with limited resources from data streams prone to severe distribution shifts. 
The main body of works presumes the presence of tasks, which divide the data streams into large discrete subsets, and are indicated to the learner with a task identifier~\cite{kirkpatrick2017overcoming,li2017learning,zenke2017continual,aljundi2018memory,lange2020unsupervised}.
Replay methods retain representative data for observed data distributions, currently unavailable in the learner's horizon $\mathcal{D}$.
The replay data is either obtained directly from operational memory $\mathcal{M}$ with stored samples \cite{rebuffi2017icarl,lopez2017gradient} or generated using generative models \cite{shin2017continual,kamra2017deep,seff2017continual,wu2018incremental}.
\textbf{GEM}~\cite{lopez2017gradient} uses replay in a constraint optimization perspective to project gradients towards a local joint task optimum. 
\textbf{iCaRL}~\cite{rebuffi2017icarl} employs exemplars to distill knowledge~\cite{hinton2015distilling} to the learner from a previous model version, with new class exemplars stored in a queue to optimally represent the class mean in feature space. 
The prototypes are then used for nearest neighbor prediction by the evaluator, in the same vein as 
\cite{han2020continual}.
Nonetheless, all three works strongly rely on task identifier $t$ for the learner, mostly unavailable for real-world data streams.
Moreover, in both prototypical approaches~\cite{rebuffi2017icarl,han2020continual} the prototypes remain static between the given task transitions and become outdated. Consequently, before using the evaluator they 
exhaustively recalculate the prototypes based on all exemplars in memory. 
\emph{
In contrast, our prototypes evolve in an online fashion with the data stream and remain representative for the continual learner and evaluator at all times. }

Recent works focus on online data incremental learning (Section \ref{sec:framework}) in which the learner operates completely task-free. \textbf{Reservoir}~\cite{vitter1985random} is a replay baseline with strong potential to outperform continual learning methods \cite{chaudhry2019continual}. Samples are stored in replay memory $\mathcal{M}_r$ with probability $|\mathcal{M}_r|/n$, with $n$ the number of observed samples and fixed replay buffer size $|\mathcal{M}_r|$.
\textbf{MIR}~\cite{aljundi2019online} extends Reservoir sampling with a loss-based retrieval strategy, with the cost of additional forward passes and a model copy to attain the losses for a subset of samples. 
The Reservoir buffer population approximately follows the data stream distribution, severely deteriorating the performance of underrepresented tasks in imbalanced data streams, as shown in Section~\ref{sec:unablanced-exps}.
An alternative memory population scheme is used in \textbf{GSS}~\cite{aljundi2019gradient} by extending the GEM constraint optimization perspective to an instance-based level. GSS adds samples to the buffer based on their gradients, whereas GEM requires the number of tasks and the task transitions to divide memory equally over all tasks a priori. 
Another memory population is used in iCaRL, incrementally subdividing over all classes after learning a task by iteratively adding observed samples from the horizon $\mathcal{D}_t$ to optimally approximate the class mean in feature space.
As this is computationally expensive,
recent works
explore other balancing schemes~\cite{kim2020imbalanced,chrysakis2020online}, where we propose a simple but effective class-based Reservoir scheme with uniform retrieval.

Another branch of works are parameter isolation methods~\cite{de2019continual}, allocating parameters to subsets of the data. Several task incremental works assign parameters based on the task identifier \cite{mallya2018packnet,serra2018overcoming}.
A new line of work instead focuses on data incremental model expansion. \textbf{CURL}~\cite{rao2019continual} enables task-free and unsupervised adaptation using a multi-component variational auto-encoder, with generative replay from a model copy avoiding forgetting in the current model.
\textbf{CN-DPM}~\cite{lee2020neural} allocates data subsets to expert networks following a Dirichlet process mixture.
In contrast to these capacity expansion based methods, CoPE evades unbound allocation of resources, as the memory and network capacity are fixed with the replay memory dynamically subdivided over categories occurring in the data stream. 
Note that new categories require an additional prototype, but these are only $d$-dimensional and therefore insignificant in size, and the set of categories is typically limited as well.

\begin{figure*}[!ht]
  \centering
  \includegraphics[clip,trim={0cm 0cm 0cm 0.2cm},width=0.9\linewidth]{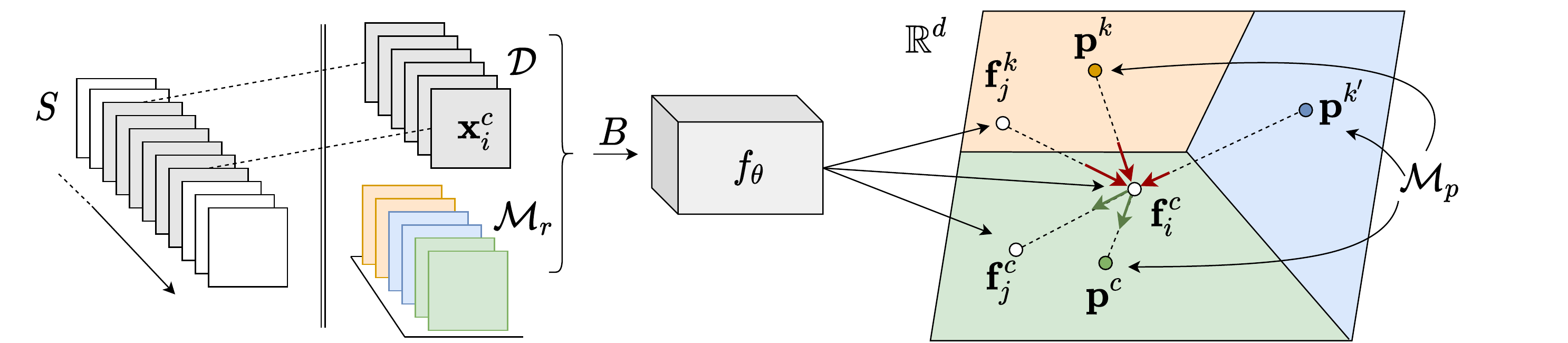}
    \caption{\label{fig:mainsetup}
       Main setup. The learner updates network $f_\theta$ and prototypes ${\bf p}^y, \forall y \in \mathcal{Y}$ continually.
       The PPP-loss encourages inter-class variance (red arrows) and reduces intra-class variance (green arrows).
}
\vspace{-0.15cm}
\end{figure*}

Besides the focus on continual learning in this work, our learner-evaluator framework generalizes to concept drift as well~\cite{schlimmer1986beyond}, for which we refer to an overview in \cite{tsymbal2004problem,gama2014survey}.
Further, in deep embedding learning most commonly pairs \cite{hadsell2006dimensionality} and triplets~\cite{harwood2017smart} of samples are considered in contrastive losses, whereas other works use batch information in lifted structure embeddings~\cite{oh2016deep} or
instance-wise softmax embeddings \cite{ye2019unsupervised}.
These approaches fully depend on the batch size, whereas our pseudo-prototypical proxy loss aggregates both decoupled prototypes and the additional batch pseudo-prototypes to defy class interference in the latent space. 
Learning prototypical representations also shows promising results in few-shot learning ~\cite{snell2017prototypical}.

\section{Continual Prototype Evolution}
\label{sec:method}

The online data incremental learning setup of the CoPE learner is described in Figure~\ref{fig:mainsetup}. Embedding network $f_\theta$ maps processing batch $B$, composed of samples in horizon $\mathcal{D}$ from the non-iid data stream $S$ and operational memory $\mathcal{M}$, to low-dimensional $\mathbb{R}^d$ latent space, followed by a nearest neighbor classifier.  We enforce $||f_{\theta}({\bf x}_i)||=1$ with an L2 normalization layer.
$\mathcal{M}$ is subdivided in replay memory $\mathcal{M}_r$ and prototypical memory $\mathcal{M}_p$.
CoPE comprises three main components: continually evolving representations, balanced replay and the pseudo-prototypical proxy (PPP) loss.
In the following, we discuss these components and formalize the optimal choice of prototype, with ${\bf f}_i^c$ denoting latent space projection $f_{\theta}({\bf x}_i^c)$ for an instance ${\bf x}_i$ of class $c$. For the full algorithm, we refer to Appendix~A.

\subsection{Evolving representations}

Each observed class $c \in \mathcal{Y}$ is represented by a slowly progressing prototype $\bf p^c$ in operational memory $\mathcal{M}_p$. The nearest neighbor classifier finds the most similar prototype for the given query ${\bf x}_i$, predicting $c^* = \argmax_{c \in \mathcal{Y}} \bf f_i^T  p^c$.
Similar to \cite{mensink2013distance, rebuffi2017icarl}, the class-prototype approximates the center of mass in the latent space, which we formally justify in Section~\ref{sec:optimal}.
The main crux with storing representations is to prevent them from becoming obsolete as the embedding network evolves.
This is further complicated by the shifting data distributions in the non-stationary regime, incurring catastrophic forgetting. 
Experience replay from a buffer $\mathcal{M}_r$ is a well known approach to address this forgetting.
In our setup the replayed exemplars provide additional information about the current state of the embedding space, enabling rehearsal to rectify approximation ${\bf p}^c$ to the true center of mass.
Concretely, the sampled batch $B_n$ equals the horizon $\mathcal{D}$ from data stream $S$ and joins batch $B_\mathcal{M}$ of equal size from memory $\mathcal{M}_r$, constituting $B$ as $B_n \cup B_\mathcal{M}$.
However, updating the prototypes by fully relying on features extracted from $B$ incurs an unstable optimization process as the representative prototypes depend on stochastic sampling of the class distributions.
Therefore, we design the prototypes to evolve continually with a high momentum based update for each observed batch, aiming to stabilize the impetuous changes in the data stream:
\begin{equation}
    \label{eq:p}
    \begin{split}
    {\bf  p}^c \leftarrow& \alpha {\bf  p}^c + (1 - \alpha) {\bf \Bar{p}}^c, \\
    &\text{s.t. } \ {\bf \Bar{p}}^c = \frac{1}{|B^c|} \sum_{{\bf x}^c \in B^c}f_\theta({\bf x}^c)
    \end{split}
\end{equation}
with momentum parameter $\alpha \in [0,1]$, the batch subset  $B^c = \{({\bf x}_i, y_i=c) \in B \}$ of class $c$, and ${\bf \Bar{p}}^c$ the corresponding center of mass in latent space for the current batch. 
Due to triangle inequality ${\bf p}^c$ is no longer unit length and requires to be L2-normalized after the update in \Eqref{eq:p}.
We empirically validate the effectiveness of high momentum with $\alpha \approx 1$ in the ablation study in Appendix~D.

\subsection{Balanced replay}

Similar to \cite{rebuffi2017icarl, chrysakis2020online}, the total buffer size $|\mathcal{M}_r|$ is equally divided over the number of observed classes $|\mathcal{Y}|$ in a dynamic fashion. This scheme ensures consistent buffer capacity for all classes, making memory allocation independent of the data stream characteristics. 
As $S$ is typically highly imbalanced in real-world scenarios, this memory scheme prevents classes to be eradicated from the buffer and assumes equal importance to represent each class at all times.
Consequently, random retrieval from the buffer resembles sampling an iid replay batch. 
Furthermore, each class-specific replay memory $\mathcal{M}^c_r$ can simply capture a random subset of its parent class distribution to approximate its center of mass.
This avoids computationally expensive herding techniques as in iCaRL \cite{rebuffi2017icarl}, which would require recalculation of the feature means for each change of the memory size or network parameters.

\subsection{Pseudo-Prototypical Proxy loss}
The learner optimizes $f_{\theta}$ to project an instance ${{\bf f}_i^c \in \mathbb{R}^d}$ of class $c$ close to its corresponding prototype ${\bf p}^c$ in the latent space.
As the prototype acts as a surrogate for the class mean in latent space, the cluster population has a common reference point to reduce intra-class variance, and we can enforce inter-class variance by remaining distant from the other class prototypes.
Additionally, due to the embedding architecture we can use intrinsic information of the batch samples in the latent space.
Therefore, we exploit the supervision signal ${\bf y}_i$ in a sample $({\bf x}_i,{\bf y}_i) \in B$ not only to indicate which class ${\bf x}_i$ belongs to, but also to make the distinction between positive and negative pairs in $B$.
Consequently, we can define one-against-all subsets  for an instance of class $c$, with positives from the same class in $B^c = \{({\bf x}_i, y_i=c) \in B \}$ and negatives in $B^k$.
Starting from these sets, the prototypical attractor and repellor sets for an instance ${\bf x}_i^c$ are constituted with the class prototype ${\bf p}^c$ and the other instances in $B$.
First, the other instances of class $c$ act as pseudo-prototypes $\hat{\bf p}^c$ in attractor set  $\mathbb{P}^c_{i} =  \{{\bf p}^c\} \cup \{\hat{\bf p}^c_j = f_{\theta}({\bf x}_j^c) \ |\  \forall {\bf x}_j^c \in B^c,\ i \ne j\}$.
Second, the samples of other classes ${\bf x}_j^k \in B^k$ should instead avoid both ${\bf x}_i^c$ in latent space and the class representative ${\bf p}^c$, defined by repellor set $\mathbb{U}^c_{i} =  \{ {\bf p}^c, \ \hat{\bf p}^c_i = f_{\theta}({\bf x}_i^c) \}$.
The attractor set $\mathbb{P}^c_{i}$ for ${\bf x}_i^c$ facilitates a decrease in intra-class variance with ${\bf p}^c$ safeguarding the absence of positive batch pairs with ${1 \leq |\mathbb{P}^c_{i}| \leq |B^c|}$, whereas the repellor $\mathbb{U}^c_{i}$ exploits ${\bf x}_i^c$ and corresponding prototype as a reference point to increase inter-class variance.
To incorporate the attractor and repellor sets, we formulate a binary classification problem similar to \cite{ye2019unsupervised}, with the joint probability that instance ${\bf x}_i^c$ is predicted as class $c$ and instances ${\bf x}_j^k \in B^k$ not being predicted as class $c$ in 
\begin{equation}
    \label{eq:prob_joint}
    P_i = P(c |{\bf x}_i^c)\prod_{{\bf x}_j^k} (1- P_i(c|{\bf x}_j^k))
\end{equation}
with the assumption of independence between  ${\bf x}_i^c$ and ${\bf x}_j^k$ being recognized as $c$.
We define the expected posterior probabilities for the attractor and repellor sets of instance ${\bf x}_i^c$ respectively as
\begin{equation}
    \label{eq:pos_exp1}
        P(c|{\bf x}_i^c) =  \mathop{\mathbb{E}}_{\Tilde{\bf p}^c \in  \mathbb{P}^c_{i}} \left[ P(c|{\bf f}_i^c, \Tilde{\bf p}^c)\right]
\end{equation}
\begin{equation}
    \label{eq:pos_exp2}
    P_i(c|{\bf x}_j^k) = \mathop{\mathbb{E}}_{\Tilde{\bf p}^c \in  \mathbb{U}^c_{i}} \left[ P(c|{\bf f}_j^k, \Tilde{\bf p}^c)\right] 
\end{equation}
 with $\Tilde{\bf p}^c$ a proxy for the latent mean of class $c$ in
\begin{equation} \label{eq:probc}
    P(c|{\bf f}, \Tilde{\bf p}^c) = \frac{\exp({\bf f}^{T} \Tilde{\bf p}^c/\tau)}{\exp({\bf f}^T \Tilde{\bf p}^c/\tau) + \sum_{k \ne c}\exp({\bf f}^T {\bf p}^{k}/\tau)} 
\end{equation}
where temperature $\tau$ controls the concentration level of the distribution \cite{hinton2015distilling}, assuming a cosine similarity metric ${\bf f}_i^{T} {\bf f}_j$ with vectors normalized to unit length.
We reformulate the objective in \Eqref{eq:prob_joint} as loss function $\mathcal{L}$
by taking the negative log-likelihood and summing over all the instances in $B$, which approximates the true joint probability with assumed independent pairs in the batch:
\begin{equation}
    \label{eq:final_loss}
    \begin{split}
    \mathcal{L} = - \frac{1}{|B|} \bigg[ & \sum_i \log P(c |{\bf x}_i^c)  \\
     &+ \sum_i\sum_{{\bf x}_j^k} \log(1- P_i(c|{\bf x}_j^k)) \bigg] .
    \end{split}
\end{equation}

\subsection{Optimal prototypes}
\label{sec:optimal}

We update prototypes to approximate the mean of the parent distribution in  \Eqref{eq:p}. This assumption is optimal for Bregman divergences for which the cluster mean is shown to have minimal distance to its population~\cite{banerjee2005clustering}.
This Bregman divergence is defined for a differentiable, strictly convex function $\varphi$ as
\begin{equation}
    d_{\varphi}({\bf f}_i,{\bf f}_j) = \varphi ({\bf f}_i) - \varphi({\bf f}_j) - ({\bf f}_i - {\bf f}_j)^T \nabla \varphi ({\bf f}_j),
\end{equation}
for which the squared Euclidean distance with $\varphi({\bf f}) = ||{\bf f}||^2$ is a canonical example. 
The squared Euclidean distance is proportional to the cosine distance with vectors normalized to unit length: 
${\frac{1}{2}||{\bf f}_i - {\bf f}_j||^2 = 1 - \cos \angle({\bf f}_i, {\bf f}_j)}$.
As the PPP-loss for \Eqref{eq:probc} requires a similarity measure instead of a distance measure, we employ the complementary normalized cosine similarity $\cos \angle({\bf f}_i, {\bf f}_j) = {\bf f}_i^{T} {\bf f}_j$ with  $||{\bf f}_i|| = || {\bf f}_j|| = 1$. Besides the desirable cluster-mean property of its complement, this metric is also efficient for calculating the full batch similarity matrix using matrix multiplication libraries.

\section{Experiments}
 \label{sec:exps}
This work examines five balanced data streams and 15 highly imbalanced variants based on Split-MNIST, Split-CIFAR10 and Split-CIFAR100, from which two low-capacity balanced setups are discussed in Appendix~E.
The learner is presented a data stream $S$, constituted by a sequence of tasks, each delineated by a subset of classes from the original dataset.
Although the learner in CoPE is completely ignorant to the notion of task, this setup enables comparing to methods requiring task boundaries such as GEM and iCaRL.
The evaluator uses a held-out dataset of static concepts in $S_{eval}$, evaluating with the subset of seen concepts $\mathcal{Y}$ in $\mathcal{D}_{eval}$ using the accuracy metric.
The CoPE learner processes data online with $B_n=\mathcal{D}$ in the data incremental setup. To enable fair comparison, this allows per-task processing of 1 epoch for methods requiring task boundaries with $B \subset \mathcal{D}_t$. We use vanilla stochastic gradient descent with a limited processing batch size $|B_n|$ of 10 as in~\cite{lopez2017gradient,aljundi2019gradient,lee2020neural}.
All results are averaged over 5 different network initializations.
Appendix details the full setup with additional experiments.

\par
\textbf{Balanced data streams} contain a similar amount of data per task. We consider three benchmarks.
First, \textbf{Split-MNIST}  constitutes the MNIST~\cite{lecun1998gradient} handwritten digit recognition dataset with 60k training samples, split into 5 tasks according to pairs of incrementing digits. 
Second, \textbf{Split-CIFAR10} considers the CIFAR10~\cite{krizhevsky2009learning} dataset, subdivided into 5 tasks with 2 labels each, where each task entails 10k training samples.
Third, \textbf{Split-CIFAR100} is a variant of the CIFAR dataset with 100 different classes. The 50k training samples are subdivided in 20 tasks of 2.5k samples as in \cite{lopez2017gradient, lee2020neural}. For all datasets the evaluator considers the entire original test subset for $S_{eval}$.

\par
\textbf{Imbalanced data streams} introduce a more realistic scenario without equality assumptions on the task durations in $S$. This addresses a common weakness in the literature mostly balancing the data streams artificially.
Besides the imbalanced Split-MNIST setup \cite{aljundi2019gradient}, we introduce two novel and more challenging benchmarks based on Split-CIFAR10 and Split-CIFAR100, where data stream $S$ comprises significantly more data in task $T_i$, denoted by $S(T_i)$.
Split-MNIST and Split-CIFAR10 have respectively 2k and 4k samples in  $T_i$, whereas tasks $T_j$ for $j \ne i$ contain factor $10$ less data for five variants $S(T_i), \ \forall i \in \{1,...,5 \}$. Split-CIFAR100 defines $T_i$ with 2.5k samples and 1k for the remaining tasks, with variants $i \in \{1,5,10,15,20 \}$.

\boldspacepar{Architectures.} MNIST setups use an MLP with 2 hidden layers of  400 units with  $2$k memories for the balanced setup as in \cite{hsu2018re, lee2020neural, van2019three}, and 100 units with $|\mathcal{M}_r|=0.3$k for the imbalanced setup as in \cite{aljundi2019gradient}.
CIFAR setups use a slim version of Resnet18~\cite{he2016deep} with a $1$k memory size for CIFAR10 \cite{aljundi2019gradient,lee2020neural}, and 5k for CIFAR100 \cite{lopez2017gradient}.%

\begin{figure}[!hbtp]
\centering
\begin{subfigure}{\linewidth}
  \centering
  \caption{Split-MNIST} \vspace{-0.2cm}
    \includegraphics[clip,trim={0.2cm -0.3cm 0.2cm 0.1cm},width=1\linewidth]{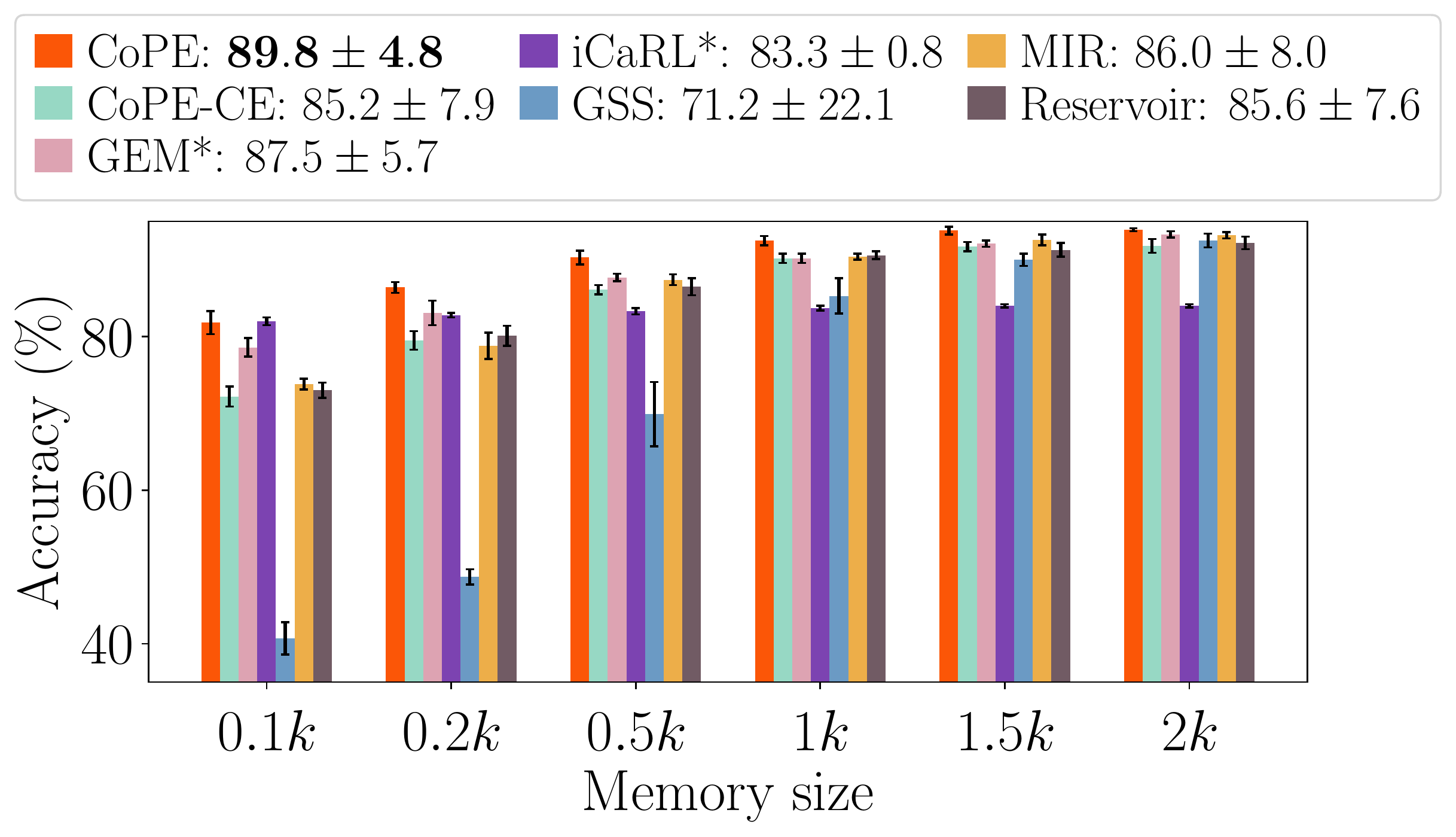}
\end{subfigure}
\begin{subfigure}{\linewidth}
  \centering
       \caption{Split-CIFAR10} \vspace{-0.2cm}
    \includegraphics[clip,trim={0.2cm -0.3cm 0.2cm 0.2cm},width=1\linewidth]{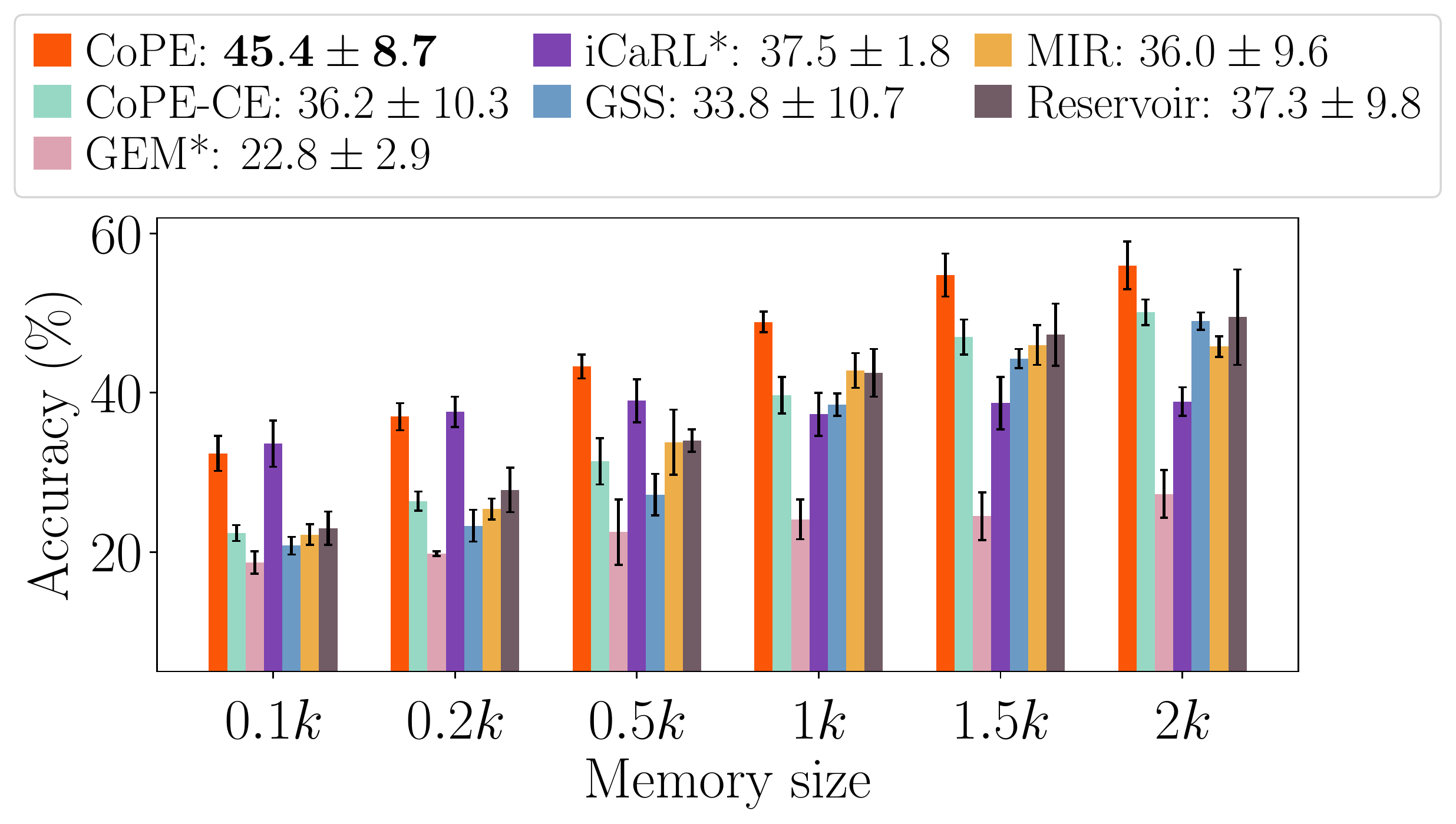}
\end{subfigure}
\begin{subfigure}{\linewidth}
  \centering
  \caption{Split-CIFAR100} \vspace{-0.2cm}
    \includegraphics[clip,trim={0.2cm 0.25cm 0.2cm 0.2cm},width=1\linewidth]{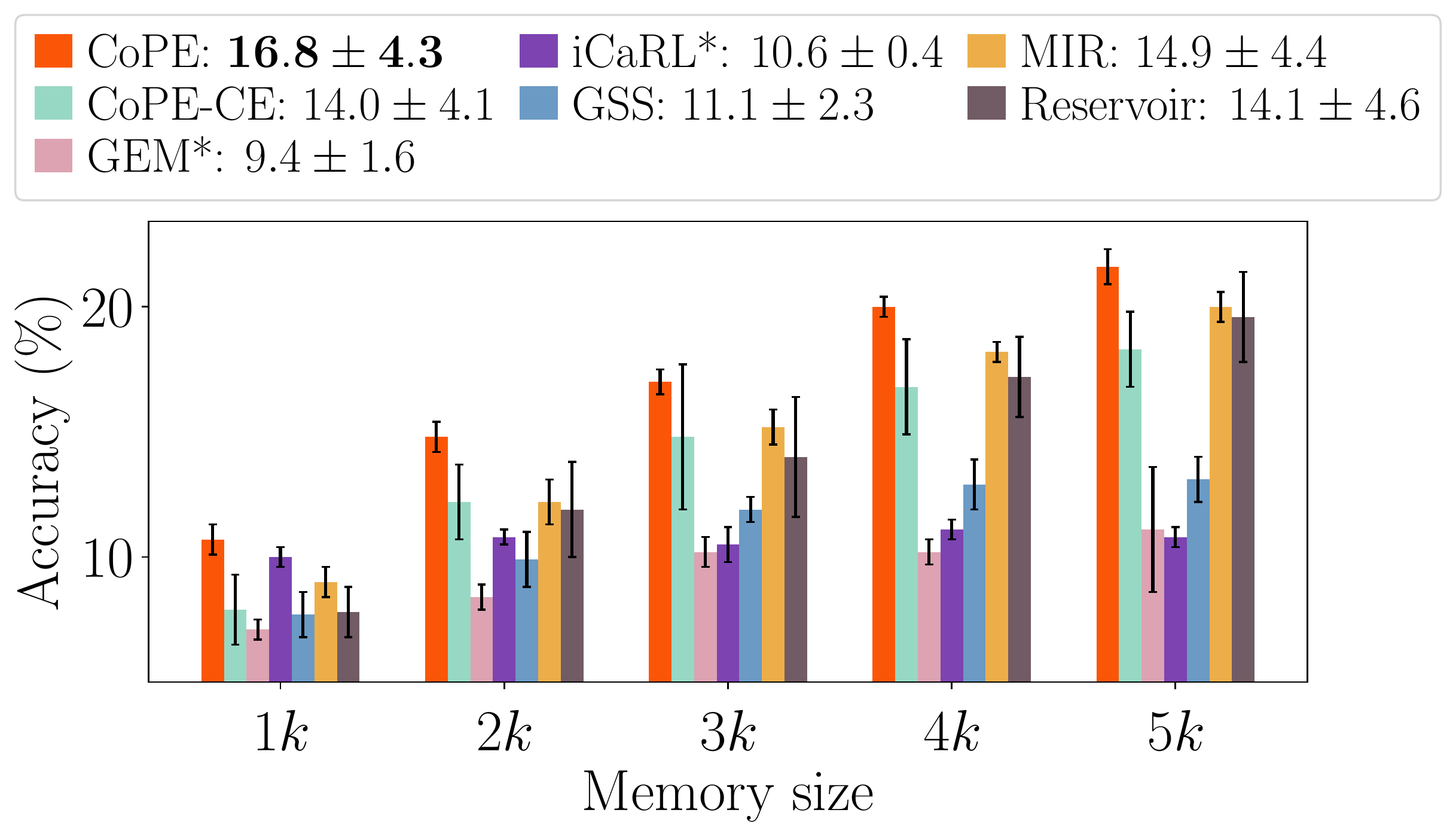}
\end{subfigure}
\caption{\label{fig:unbal}Accuracies over buffer sizes $|\mathcal{M}_r|$ for balanced Split-MNIST (\emph{top}), Split-CIFAR10 (\emph{center}) and Split-CIFAR100 (\emph{bottom}) sequences. The legend reports averages over all observed buffer sizes. 
'$*$' indicates learner using task information. %
}
\label{fig:bufferablation}
\end{figure}

\begin{figure*}[!ht]
\centering
    \begin{subfigure}{0.33\linewidth}
      \centering
        \includegraphics[clip,trim={0.3cm 0.4cm 0cm 0.2cm},width=\textwidth]{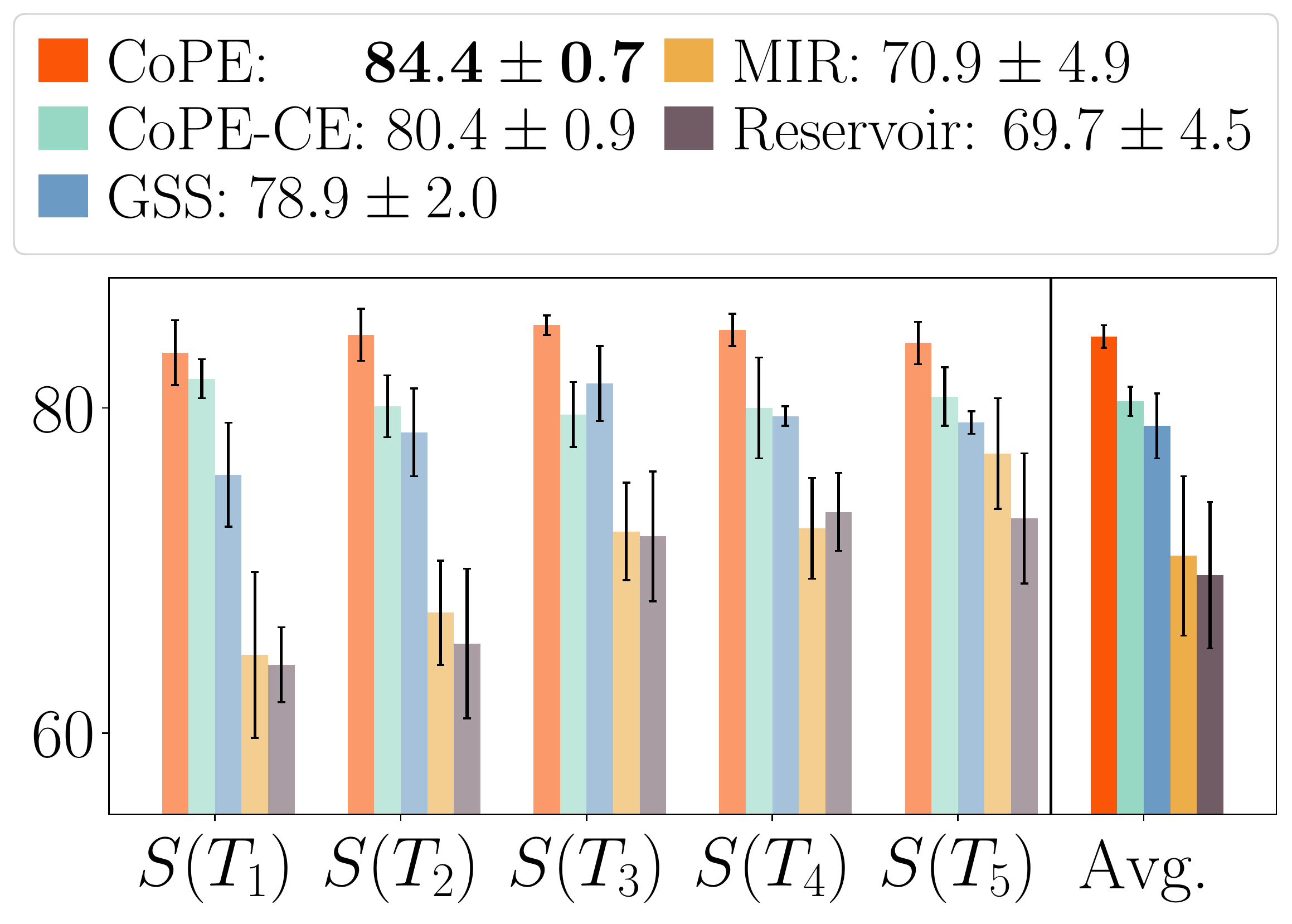}
    \end{subfigure}%
        \begin{subfigure}{0.33\linewidth}
      \centering
\includegraphics[clip,trim={0.3cm 0.4cm 0.2cm 0.2cm},width=\textwidth]{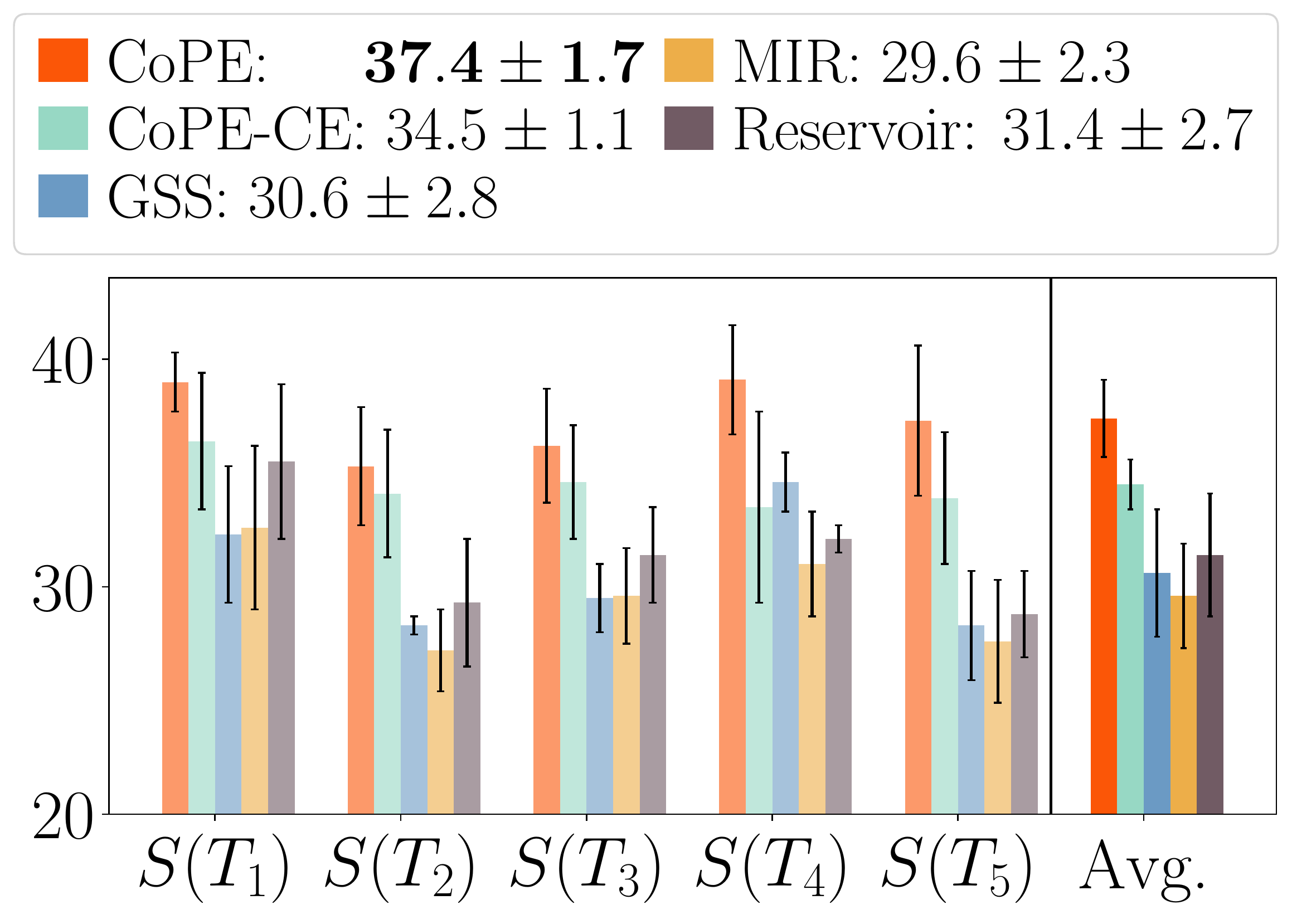}
    \end{subfigure}%
        \begin{subfigure}{0.33\linewidth}
      \centering
\includegraphics[clip,trim={0.3cm 0.4cm 0.2cm 0.2cm},width=\textwidth]{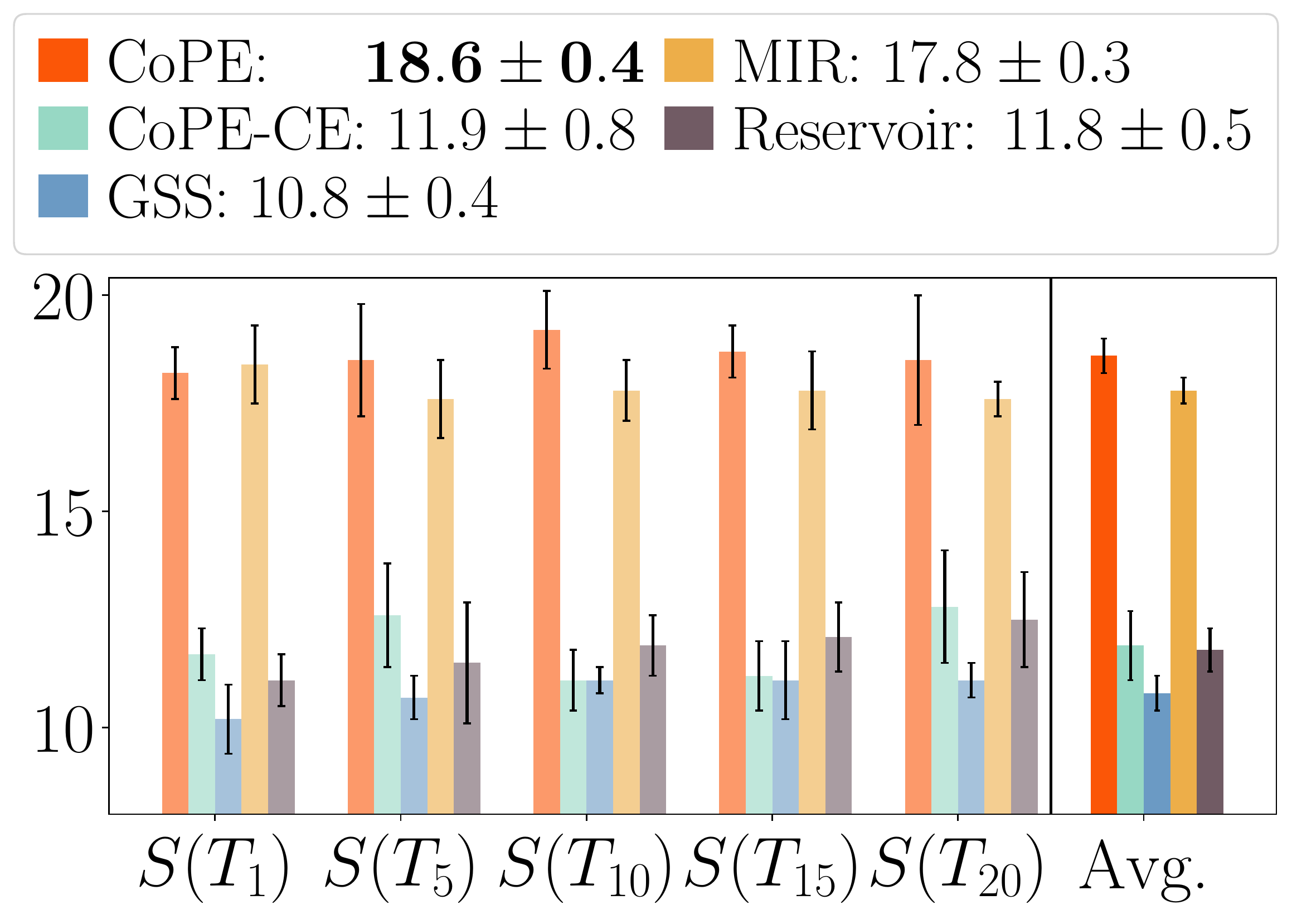}

    \end{subfigure}%
\caption{Accuracy (\%) for imbalanced Split-MNIST (\emph{left}), Split-CIFAR10 (\emph{center}) and Split-CIFAR100 (\emph{right}) sequences. The legend reports average accuracies over all the sequence variations.}
\label{fig:imbalanced-exps}
\end{figure*}

\boldspacepar{Methods} compared to CoPE entail 11 baselines, with details on prior work discussed in Section~\ref{sec:relwork}. 
The upper reference point for performance when relaxing the challenging non-iid property of continual learning is set by \textbf{iid-online \& iid-offline}. The learner shuffles the full data stream $S$ to ensure the iid property, for which \emph{iid-online} trains a single epoch and \emph{iid-offline} multiple epochs.
In contrast, the \textbf{finetune} learner considers non-iid data stream $S$ sequentially, but optimizes solely for the new batch which typically results in worst-case catastrophic forgetting. 
\textbf{CoPE-CE} is a reference point for the merits of a prototypical approach by solely using the CoPE memory and sampling scheme, but with a typical cross-entropy loss and softmax classifier.
\textbf{GEM} and \textbf{iCaRL} are standard replay methods considered in a class incremental setup, with the learner requiring knowledge about task boundaries. For online data incremental learning, we consider the \textbf{reservoir}, \textbf{MIR} and greedy \textbf{GSS} replay baselines, with
\textbf{CURL} and \textbf{CN-DPM} instead relying on model expansion.

\section{Results and discussion}
\label{sec:results}

\subsection{Balanced data streams}
\label{sec:balanced-exps}

The results for the three balanced data streams in Table~\ref{tab:bal-results} consistently report state-of-the-art results for CoPE. 
The difficulty for learning online is reflected in the discrepancy of performance between iid-offline and iid-online, indicating increasing difficulty for a minimal $2\%$ discrepancy for Split-MNIST, raised to $20\%$ for Split-CIFAR10, and culminating to $30\%$ in Split-CIFAR100. 
For Split-MNIST the gap with iid-online performance is closed by $0.7\%$ compared to main competitors GEM and DN-CPM.
Furthermore, in the more challenging Split-CIFAR10 setup CoPE significantly increases the gained margin to $3.7\%$. 
In the most challenging Split-CIFAR100, CN-DPM, Reservoir and MIR are able to perform on par with the iid-online baseline, however, CoPE establishes an improvement of at least $1.5\%$ over all four baselines.

 \begin{table}[!ht]
\resizebox{1\linewidth}{!}{%
\begin{tabular}{@{}llll@{}}
\toprule
                                & \textbf{Split-MNIST}      & \textbf{Split-CIFAR10}    & \textbf{Split-CIFAR100}   \\ \midrule
iid-offline                     & $98.44 \pm 0.02$          & $83.02 \pm 0.60$          & $50.28 \pm 0.66$          \\
iid-online                      & $96.57 \pm 0.14$          & $62.31 \pm 1.67$          & $20.10 \pm 0.90$          \\ \midrule\midrule
finetune                        & $19.75 \pm 0.05$          & $18.55 \pm 0.34$          & $3.53 \pm 0.04$           \\
GEM                             & $93.25 \pm 0.36$          & $24.13 \pm 2.46$          &  $11.12 \pm 2.48$                         \\
iCARL                           & $83.95 \pm 0.21$          & $37.32 \pm 2.66$          & $10.80 \pm 0.37$          \\
CURL \cite{rao2019continual}   & $92.59 \pm 0.66$          & $-$                      & $-$                      \\
DN-CPM \cite{lee2020neural} & $93.23 \pm 0.09$          & $45.21 \pm 0.18$          & $20.10 \pm 0.12$          \\ \addlinespace
reservoir                       & $92.16 \pm 0.75$          & $42.48 \pm 3.04$          & $19.57 \pm 1.79$          \\
MIR                             & $93.20 \pm 0.36$          & $42.80 \pm 2.22$          &    $20.00 \pm 0.57$                       \\ 
GSS                             & $92.47 \pm 0.92$          & $38.45 \pm 1.41$          & $13.10 \pm 0.94$          \\ \midrule
CoPE-CE                            & $91.77 \pm 0.87$ & $39.73 \pm 2.26$ & $18.33 \pm 1.52$	 \\
CoPE (ours)                            & \textbf{$ \bf93.94 \pm 0.20$} & \textbf{$\bf48.92 \pm 1.32$} & \textbf{$\bf 21.62 \pm 0.69$} \\ \bottomrule
\end{tabular}
}
\caption{\label{tab:bal-results}The three balanced data stream accuracies ($\%$) with standard deviation over 5 initializations. 
For expansion-based methods CURL and DN-CPM we report results from their original work.}
\end{table}

Compared to balanced replay with standard cross-entropy (CoPE-CE), the prototypical approach (CoPE) proves effective with significant gains of $2.2\%$, $9.2\%$ and $3.3\%$ respectively over the three benchmarks.
Except for GEM in Split-MNIST, class incremental learning methods GEM and iCaRL are not competitive in the online setting and additionally require from the setup to reveal an identifier $t$ to the learner.
From the expansion-based methods DN-CPM is competitive, whereas CURL is more suited for unsupervised learning and lags behind.
Although Reservoir and extension MIR perform on par with iid-online for Split-CIFAR100,
the imbalanced experiments in Section~\ref{sec:unablanced-exps} show that full reservoir-based population of the buffer strongly relies on the assumption of equally sized tasks, which is unlikely to occur in real-world data streams.

\boldspacepar{Buffer size ablation study} in Figure~\ref{fig:bufferablation} shows CoPE to prevail over all sizes of replay buffer $\mathcal{M}_r$ compared to other replay methods, extending robustness to low capacity regimes. Although iCaRL shows competitive results for low capacity, CoPE scales with growing capacity leading to significantly outperforming iCaRL with $11\%$ in Split-MNIST ($2k$) and Split-CIFAR100 ($5k$), and $17\%$ in Split-CIFAR10 ($2k$).

\subsection{Imbalanced data streams}
\label{sec:unablanced-exps}
\begin{figure*}[!h]
\centering
    \begin{subfigure}{0.250\linewidth}
      \centering

        \includegraphics[clip,trim={0.2cm 0cm 0cm 0cm},width=1\textwidth]{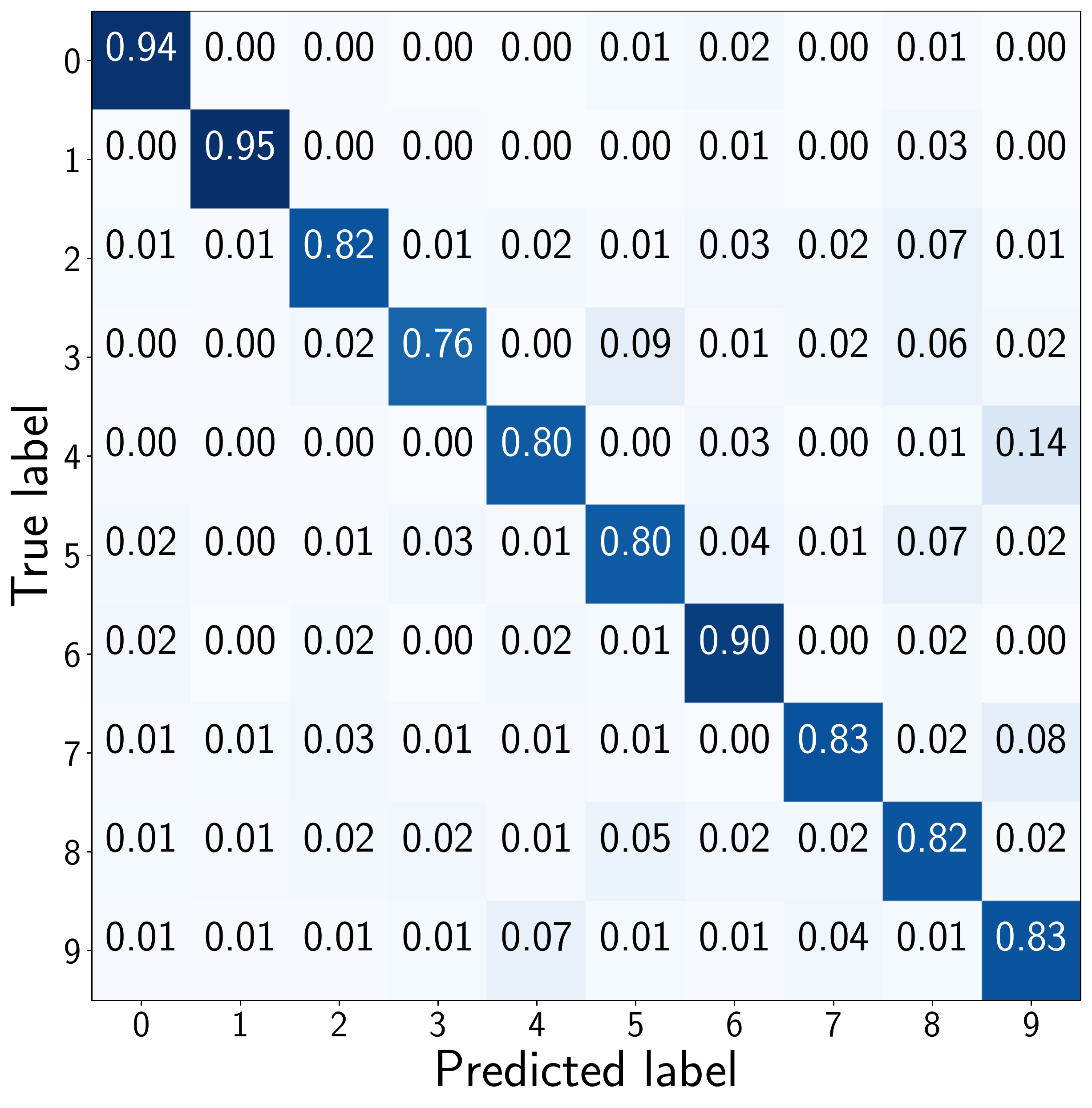} %
                \caption{CoPE (Split-MNIST)}%
    \end{subfigure}%
        \begin{subfigure}{0.24\linewidth}
      \centering

        \includegraphics[clip,trim={1.2cm 0cm 0cm 0cm},width=1\textwidth]{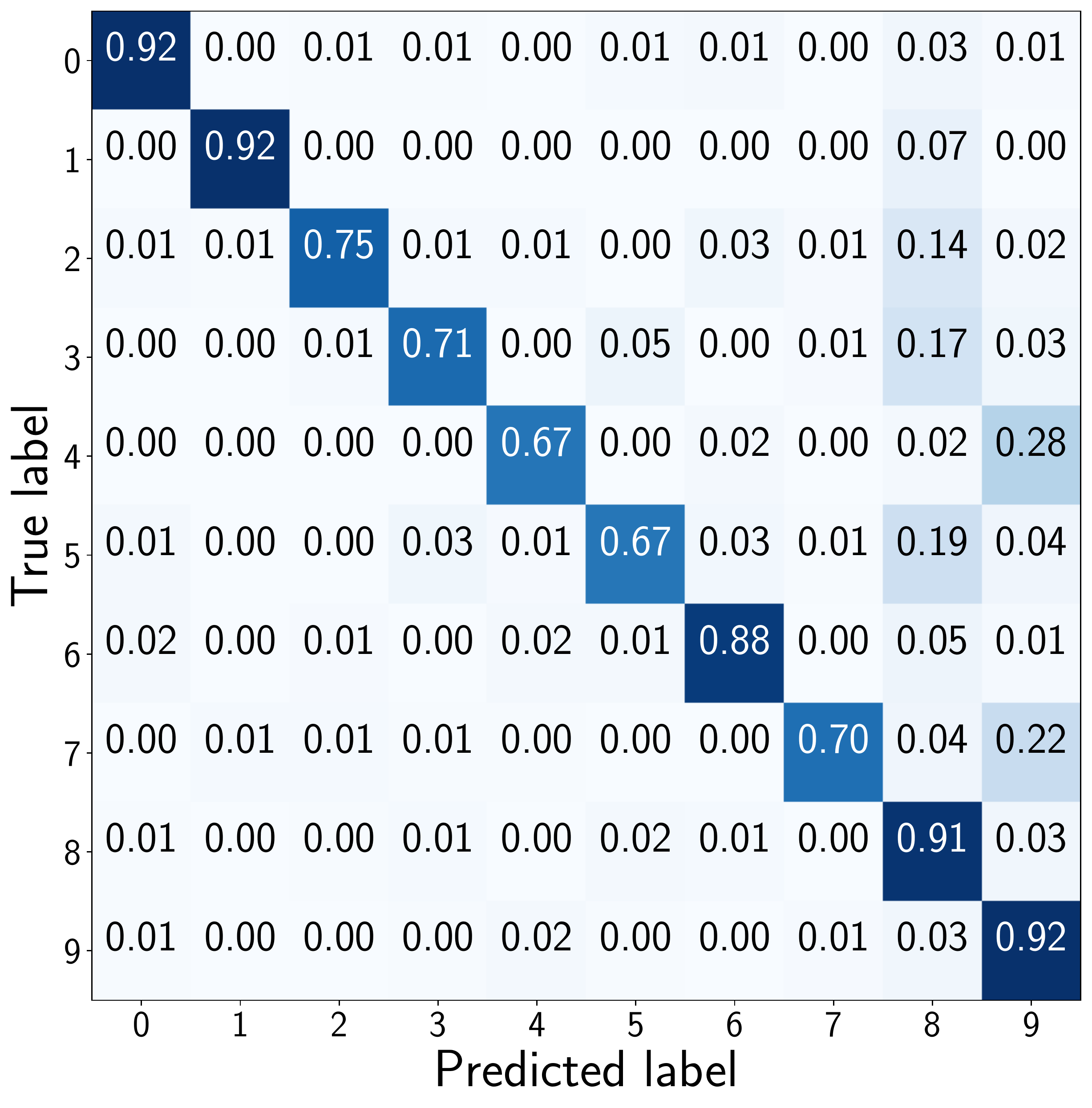} %
          \caption{CoPE-CE (Split-MNIST)}

    \end{subfigure}%
        \begin{subfigure}{0.25\linewidth}
      \centering
        \includegraphics[clip,trim={0.2cm 0cm 0cm 0cm},width=1\textwidth]{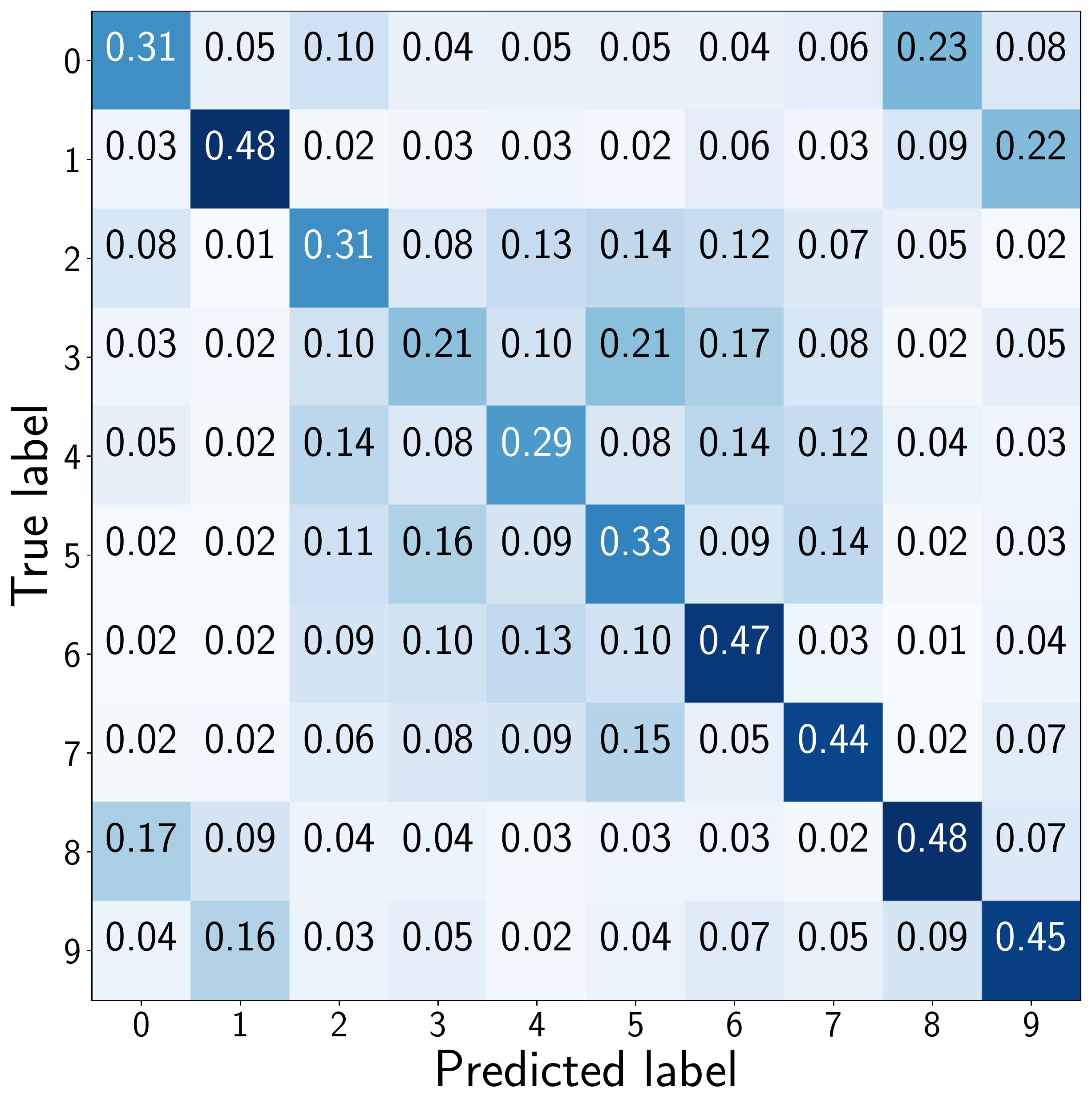} %
                    \caption{CoPE (Split-CIFAR10)}%
    \end{subfigure}%
    \begin{subfigure}{0.24\linewidth}
      \centering
        \includegraphics[clip,trim={1.2cm 0cm 0cm 0cm},width=1\textwidth]{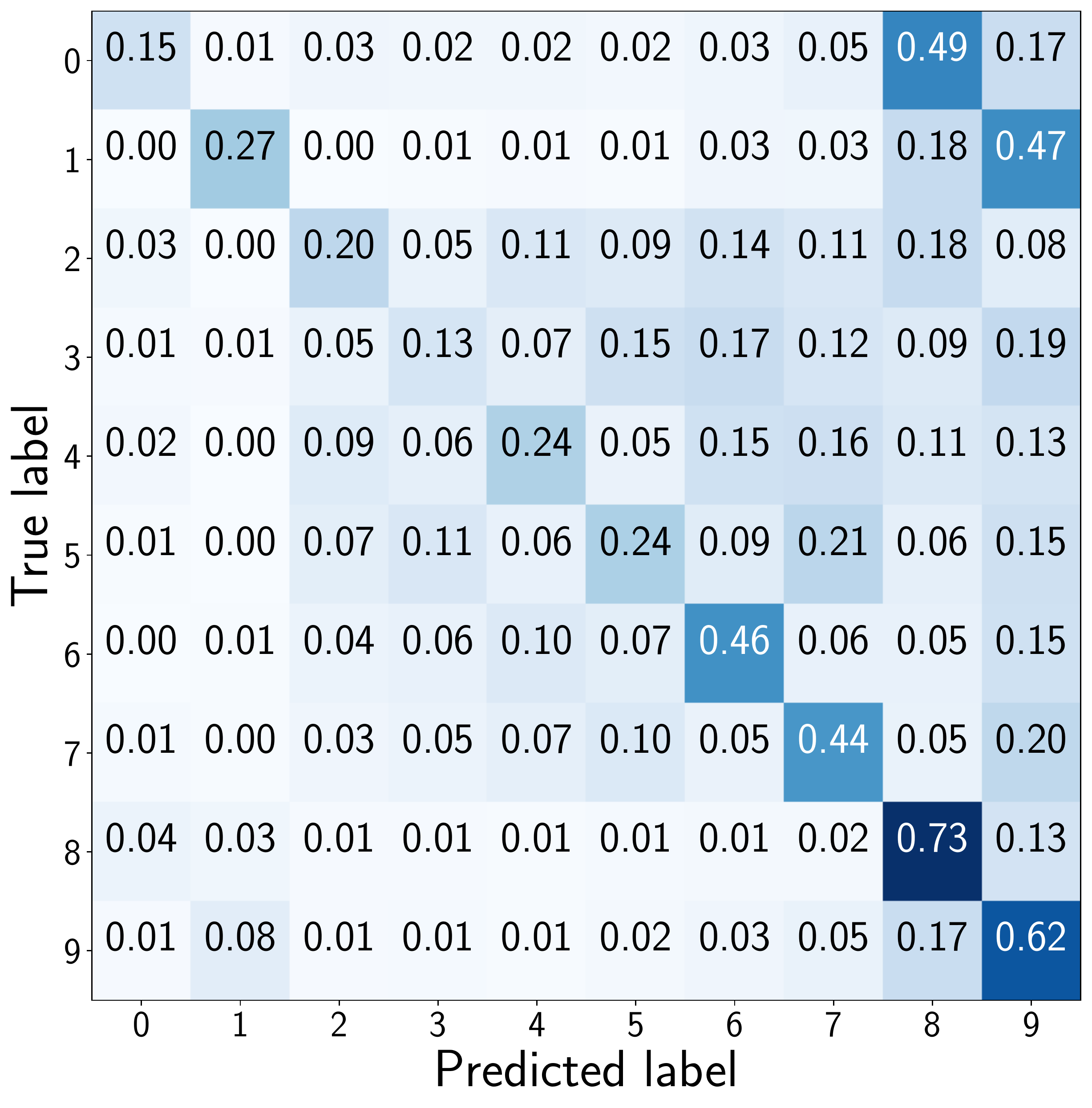} %
              \caption{CoPE-CE (Split-CIFAR10)}
    \end{subfigure}
\caption{CoPE and CoPE-CE confusion matrices at the end of learning averaged over all variations $S(T_i)$ for the imbalanced Split-MNIST setup in (a) and (b), and Split-CIFAR10 in (c) and (d).}
\label{fig:cm}
\end{figure*}

Results for the highly imbalanced data stream benchmarks are reported in Figure~\ref{fig:imbalanced-exps}.
CoPE significantly outperforms all baselines in the three scenarios, with low standard deviation for the 15 variants indicating robustness over a wide spectrum of imbalanced sequences.
GSS outperforms Reservoir and MIR for Split-MNIST, in correspondence with results in \cite{aljundi2019gradient}, whereas loss-based retrieval in MIR has significant gains for the challenging Split-CIFAR100 setting. 
However, CoPE surpasses both GSS and MIR for all three benchmarks, and on top of that operates profusely more resource efficient as discussed in Appendix~C.
The balancing memory scheme in CoPE-CE highly improves Reservoir over imbalanced Split-MNIST and Split-CIFAR10 variants with $10.8\%$ and $3.4\%$ respectively, and performs on par for Split-CIFAR100 where balancing over 100 classes with limited batch size proves more difficult.
Although CoPE and CoPE-CE share memory and retrieval schemes, the prototypical CoPE surpasses the cross-entropy based CoPE-CE with $4.0\%$, $2.9\%$ and $6.7\%$ respectively on the three benchmarks, indicating the merits of the PPP-loss and continually evolving prototypes. 
Figure~\ref{fig:cm} compares the CoPE and CoPE-CE confusion matrices at the end of learning, with CoPE-CE exhibiting high plasticity for the later observed classes, whereas CoPE better preserves the recall over early learned classes, hence effectively alleviating catastrophic forgetting.

\begin{table*}[!h]
\centering
\resizebox{0.8\textwidth}{!}{%
\begin{tabular}{@{}llllllll@{}}
\toprule
\multicolumn{1}{c}{} & \multicolumn{2}{c}{\textbf{PPP-loss}}     & \multicolumn{5}{c}{\textbf{Batch Size $|B_n|$}}                                                                                               \\ \cmidrule(lr){2-3}
\cmidrule(lr){4-8}
\multicolumn{1}{c}{} & \multicolumn{1}{c}{\textit{incl. $\hat{\bf p}$}} & \multicolumn{1}{c}{\textit{excl. $\hat{\bf p}$}} & \multicolumn{1}{c}{10 (Online)} & \multicolumn{1}{c}{20} & \multicolumn{1}{c}{50} & \multicolumn{1}{c}{100} & \multicolumn{1}{c}{200} \\ \midrule
Split-MNIST           & $93.9 \pm 0.2$           & $92.4 \pm 0.6$           & $93.9 \pm 0.2$                  & $93.9 \pm 0.6$         & $93.7 \pm 0.3$         & $93.1 \pm 0.6$          & $89.3 \pm 0.5$          \\
Split-CIFAR10              & $48.9 \pm 1.3$           & $41.3 \pm 2.0$           & $48.9 \pm 1.3$                  & $48.4 \pm 1.9$         & $43.4 \pm 2.7$         & $37.4 \pm 3.0$          & $37.0 \pm 1.3$          \\
Split-CIFAR100               & $21.6 \pm 0.7$           & $16.3 \pm 0.7$           & $21.6 \pm 0.7$                  & $21.7 \pm 0.7$         & $16.5 \pm 0.4$         & $13.8 \pm 0.5$          & $11.2 \pm 0.4$          \\ \bottomrule
\end{tabular}%
}
\caption{\label{tab:ablation:batch}Accuracies ($\%$) for ablating pseudo-prototypes $\hat{\bf p}$ in the PPP-loss and varying batch size.}
\end{table*}

\subsection{PPP-loss analysis}
\label{sec:loss-analysis}
In the challenging setting for online processing of non-iid data streams, the PPP-loss exploits information in the small processing batch $B$, introducing pseudo-prototypes $\hat{\bf p}$ on top of the prototypes. This leads to questioning to what extent the pseudo-prototypes actually contribute to the quality of the embedding, and how this relates to the batch size.
We examine both inquiries in Table~\ref{tab:ablation:batch} for the three balanced data streams by comparing inclusion and exclusion of the pseudo-prototypes $\hat{\bf p}$ in the PPP-loss, and extending the batch size $|B_n|$. 
First, including the pseudo-prototypes significantly improves overall performance, especially for the harder CIFAR-based data streams. Although both setups use batch information to update the prototypes following \Eqref{eq:p}, it seems crucial to use additional pseudo-prototypes in the PPP-loss to improve latent space quality.
Second, results for smaller batch sizes of 10 and 20 are very similar, and deteriorate towards increasing sizes.
The PPP-loss implements the expectation over the prototype and the pseudo-prototypes, assuming uniform distribution in \Eqref{eq:pos_exp1} and \Eqref{eq:pos_exp2}. Although this assumption impedes significance of the prototype for increasingly higher batch sizes, it results in ideal robustness for small online processing batches.
Small batches maintain the additional benefit of more frequent prototype and network parameter updates for the same amount of processed data. %

\section{Conclusion}
\label{sec:concl}
The novel  two-agent learner-evaluator framework introduced a new perspective on current paradigms in continual learning.
To overcome the standard paradigm of static training and testing phases, we explicitly model continual optimization and evaluation in the \emph{learner} and \emph{evaluator} agents respectively.
We formalized the required resources as the \emph{horizon} $\mathcal{D}$, containing the simultaneously available data of the data stream, and the \emph{operational memory} $\mathcal{M}$ for operation of the learning algorithm. 
Transitions in the horizon $\mathcal{D}_t \rightarrow \mathcal{D}_{t+1}$ enable a uniform differentiation between existing incremental learning paradigms
and the horizon size encloses the range from online ($\mathcal{D} = B$) to offline ($\mathcal{D} = S$) learning. 
Using the framework, we defined the task-free \emph{data incremental learning} paradigm, requiring no additional information on the horizon's identifier $t$ for both the learner and evaluator. In this challenging setup, we proposed Continual Prototype Evolution (CoPE) as a prototypical solution to learn online from non-stationary data streams. As a first, CoPE prevents the prototypes becoming obsolete in an ever evolving representation space, while using the prototypes to combat catastrophic forgetting. The three main components, continually evolving prototypes, a novel pseudo-prototypical proxy loss, and an efficient balancing replay scheme are proven remarkably effective over 11 baselines in both balanced and highly imbalanced benchmarks. 
We hope to encourage further research in online data incremental learning.


\end{document}